\documentclass{article}

\PassOptionsToPackage{numbers, compress}{natbib}
\usepackage[preprint]{neurips_2020}

\usepackage[utf8]{inputenc} % allow utf-8 input
\usepackage[T1]{fontenc}    % use 8-bit T1 fonts
\usepackage{hyperref}       % hyperlinks
\usepackage{url}            % simple URL typesetting
\usepackage{booktabs}       % professional-quality tables
\usepackage{amsfonts}       % blackboard math symbols
\usepackage{nicefrac}       % compact symbols for 1/2, etc.
\usepackage{microtype}      % microtypography
\usepackage{booktabs}       % nice tables
\usepackage{graphicx}       % images
\usepackage{subfigure}     % multiple figures together
\usepackage{float}          % floating figures
\usepackage{textcomp}
\usepackage[ruled,vlined]{algorithm2e} % pretty algorithm boxes
\graphicspath{ {./images/} }

\title{PrototypeML: A Neural Network Integrated Design and Development Environment}

\author{%
  Daniel Reiss Harris \\
  Adventurize \\
  \texttt{danielharris@alumni.harvard.edu}
}

\begin{document}

\maketitle

\begin{abstract}
    Neural network architectures are most often conceptually designed and described in visual terms, but are implemented by writing error-prone code. PrototypeML is a machine learning development environment that bridges the dichotomy between the design and development processes: it provides a highly intuitive visual neural network design interface that supports (yet abstracts) the full capabilities of the PyTorch deep learning framework, reduces model design and development time, makes debugging easier, and automates many framework and code writing idiosyncrasies. In this paper, we detail the deep learning development deficiencies that drove the implementation of PrototypeML, and propose a hybrid approach to resolve these issues without limiting network expressiveness or reducing code quality. We demonstrate the real-world benefits of a visual approach to neural network design for research, industry and teaching. Available at \small{\url{https://PrototypeML.com}}.
\end{abstract}

\section{Introduction}
    Advances in deep learning have driven significant breakthroughs in computer vision, natural language processing and dozens of other areas of computing. However, one commonly cited challenge of AI adoption in industry and research is the significant time, resources and expertise required to design and develop custom neural networks \cite{mckinsey_survey} \cite{ai_index_report}.
    
    The frameworks behind deep learning advances have progressed significantly in recent years, delivering steady improvements in neural network functionality and performance. From static compiled graph solutions such as Tensorflow \cite{tensorflow}, Caffe \cite{caffe}, CNTK \cite{cntk} and Theano \cite{theano}, to increasingly popular dynamic eager-executing frameworks such as PyTorch \cite{pytorch}, Tensorflow 2.0 \cite{tensorflow}, Chainer \cite{chainer}, DyNet \cite{dynet} and Torch \cite{torch}, deep learning frameworks implemented in Python and other programming languages have demonstrated their flexibility and capability in modeling the complex behaviors and interactions necessary for implementing neural networks. Yet, the process of expressing neural network behavior in code is both time consuming and error-prone, requiring constant reference to documentation and intricate understanding of and strict adherence to both computer science and deep learning principles. Framework wrappers such as Keras \cite{keras} and FastAI \cite{fastai} attempt to simplify the code involved in the neural network construction and training process, but do so at the expense of customization, and model expressiveness.
    
    While they are implemented in code, neural networks are most often conceptually designed and described in visual terms as an illustrated graph detailing the network layers and interactions within an architecture, which eases comprehension of network behavior. This paper introduces PrototypeML, a neural network integrated design and development tool that implements a hybrid approach to the network design process. It defines a highly intuitive visual design interface, and a standardized modular approach to neural network components that allows for fast design and development, and easy debugging of complete architectures, without sacrificing model expressiveness or performance.
    
\section{Related work}
    
    \textbf{Visual approach to design}\quad Numerous prior works have attempted to simplify various areas of the neural network design process through visual means. Tensorboard \cite{tensorflow} and Netron \cite{netron} offer non-editable visualizations of neural network models. NVIDIA Digits \cite{nvidia}, Weka \cite{lang2019wekadeeplearning4j}, DeepCognition DL Studio \cite{deepcognition}, Fabrik \cite{fabrik}, Sony Neural Network Console \cite{sony}, and IBM Watson \cite{ibm}, have all implemented some form of visual "codeless" neural network design tool with intuitive static, or drag-and-drop interfaces, and varying levels of support for neural network framework features (table 1). Prior work has recognized the potential value of a visual approach to neural network design and editing of models in terms of speed, accuracy, and level of expertise required \cite{ibm}. However by prioritizing purely "codeless" editing of networks, prior works demonstrate reduced support for model capability and expressiveness. Though these visualizations and editing tools are useful for basic neural network design, they fail to offer sufficient features and the customization required to construct most modern complex neural networks (for example, support for custom layers, modularized multi-module networks, comprehensive skip connection support, etc.).
    \\
    \\
    \textbf{Underlying neural network representation}\quad In order to construct neural networks via a visual interface, the various layers and components of a network must be represented such that they can be individually manipulated, positioned, and connected to other components to form a network. Works such as ONNX \cite{onnx} and Fabrik \cite{fabrik} implement a specification for a neural network intermediate format that defines the type of neural network layer (linear, convolution, etc.), and required parameters. Subsequently the intermediate format can be exported to various frameworks. While allowing for greater flexibility in choice of framework, this solution suffers from several significant drawbacks: First, neural networks expressed in intermediate formats are, by nature, forced to limit their functionality to the subset supported by all the frameworks to which they interface, thus making supporting future framework additions and custom layers a difficult and potentially time consuming task. Second, intermediate representations have thus far been limited to static graph frameworks, thus forfeiting the increasingly popular flexibility and debugging benefits that dynamic graph frameworks such as PyTorch \cite{pytorch} and Tensorflow 2.0 \cite{tensorflow} provide.

\begin{table}[H]
    \caption{Comparison of selected features between the most feature-complete tools and frameworks}
    \label{tab:related-work}
    \centering
    \begin{tabular}{lllllll}
        \toprule
        \cmidrule(r){1-6}
        Feature & DL Studio\cite{deepcognition} & Fabrik\cite{fabrik} & Sony\cite{sony} & IBM\cite{ibm} & PrototypeML \\
        \midrule
        Drag \& Drop GUI & \checkmark & \checkmark & \checkmark & \checkmark & \checkmark\\
        Custom Components & \checkmark & \checkmark & $\times$ & $\times$ & \checkmark \\
        Modular Networks & $\times$ & $\times$ & $\times$ & $\times$ & \checkmark \\
        Dynamic Graph Support & $\times$ & $\times$ & $\times$ & $\times$ & \checkmark\\
        Component Repository & $\times$ & $\times$ & $\times$ & $\times$  & \checkmark \\
        Code Importing & \checkmark & $\times$ & \checkmark & $\times$ & \checkmark\\
        Code Exporting & \checkmark & $\times$ & \checkmark & \checkmark & \checkmark\\
        Model Validation & \checkmark & $\times$ & \checkmark & \checkmark & \checkmark\\
        Skip Connections & $\times$ & $\times$ & $\times$ & $\times$ & \checkmark\\
        Multiple Inputs \& Outputs & $\times$ & $\times$ & $\times$ & $\times$ & \checkmark\\
        Complex Parameters & $\times$ & $\times$ & $\times$ & $\times$ & \checkmark\\
        Parameter Passing & $\times$ & $\times$ & $\times$ & $\times$ & \checkmark\\
        Multi-framework Support & \checkmark & \checkmark & \checkmark & \checkmark & Via ONNX\\
        Data Management & \checkmark & $\times$ & \checkmark & \checkmark & Future work\\
        Model Training & \checkmark & $\times$ & \checkmark & \checkmark & Future work\\
        \bottomrule
    \end{tabular}
\end{table}

\begin{figure}[H]
    \centering
    \subfigure[Project components]{{\includegraphics[width=0.205\textwidth, height=250pt]{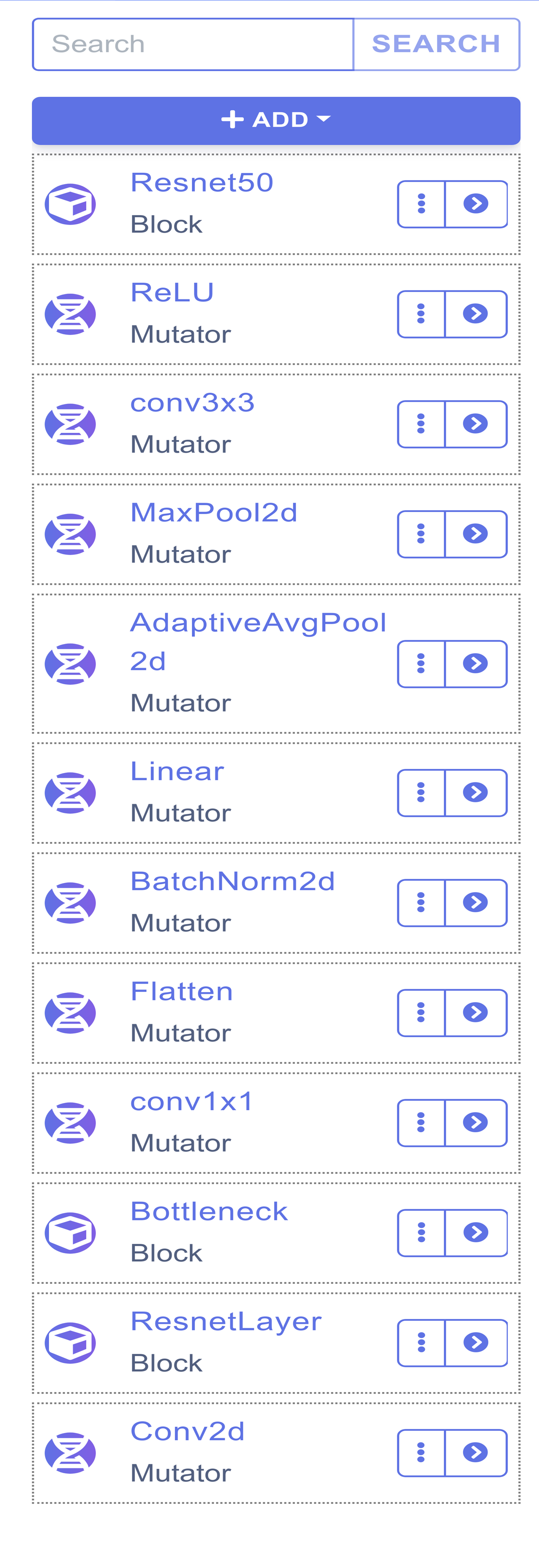} }\label{interface1}}%
    \subfigure[Drag \& drop block editor interface]{{\includegraphics[width=0.53\textwidth, height=250pt]{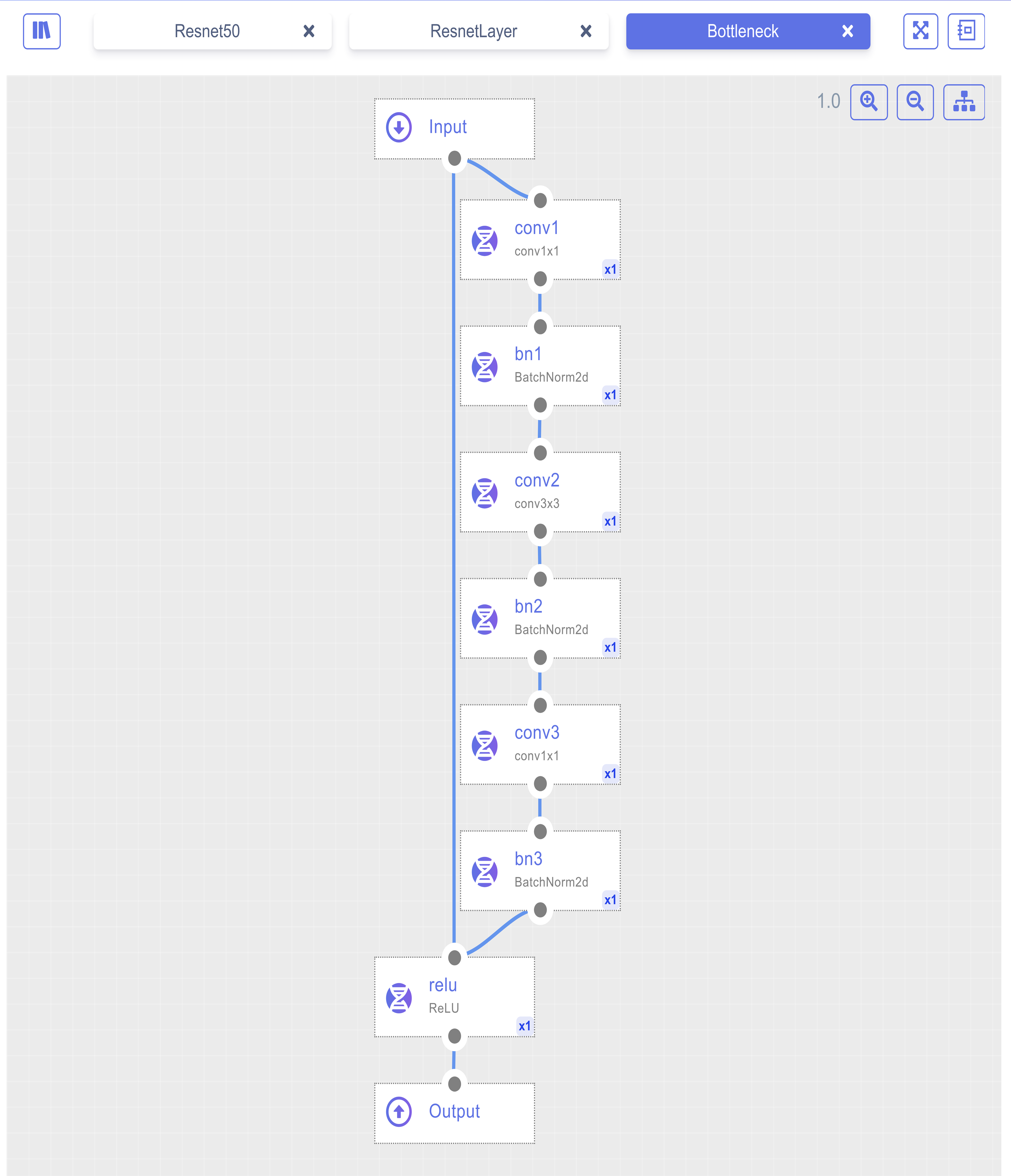} }\label{interface2}}%
    \subfigure[Parameter editor]{{\includegraphics[width=0.25\textwidth, height=250pt]{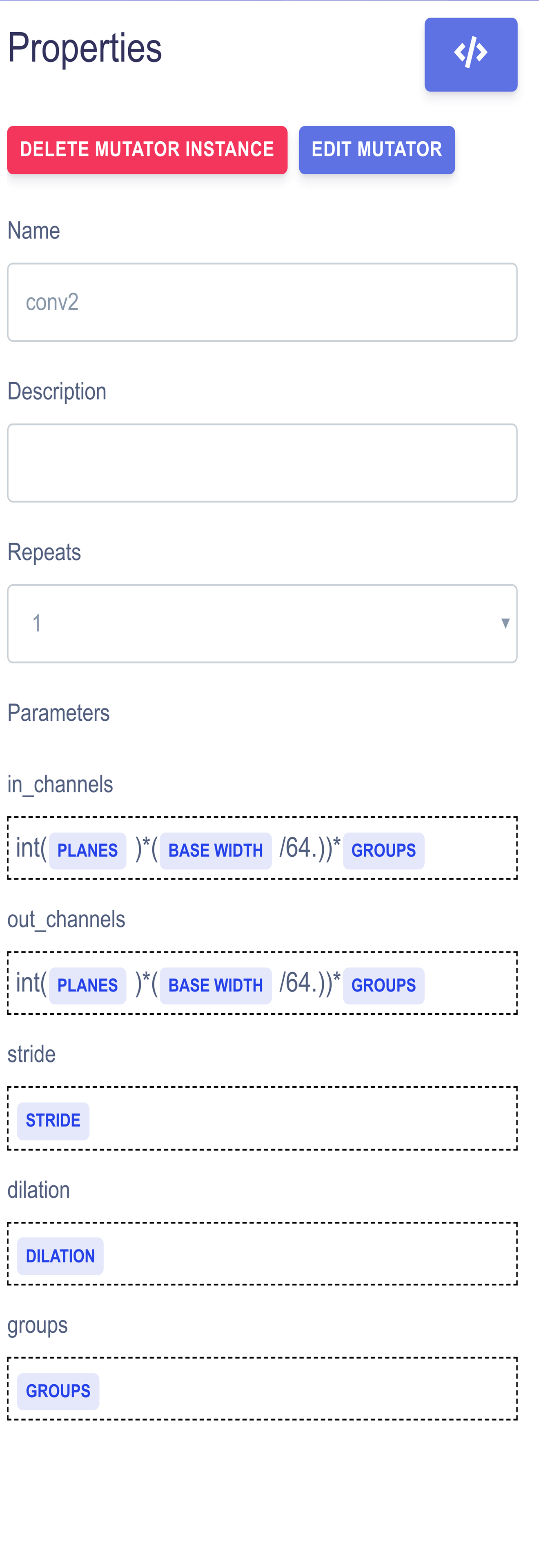}}\label{interface3}}%
    \caption{The PrototypeML interface is split into three primary sections}%
    \label{fig:interface}%
\end{figure}

\section{Proposed approach}
    PrototypeML approaches implementing a comprehensive neural network design and development environment from an entirely new direction to avoid the difficulties inherent in the intermediate format and "codeless" design paradigm employed by prior works. In this section we detail how we resolve these challenges, and support a far wider array of functionality, customization, and ability to quickly adapt to new deep learning developments.
    
    \subsection{Design principles}
        \textbf{Complete framework support}\quad We tightly couple the PrototypeML system with the PyTorch framework \cite{pytorch}, rather than attempt to offer only partial support for a variety of frameworks. We chose PyTorch for its conceptual simplicity, dynamic graph support, modular and future-proof design, and its increasing popularity with the research community. Limiting PrototypeML to PyTorch allowed us to concentrate on a single design paradigm, implement the full range of PyTorch functionality, and take advantage of the existing ecosystem and previously-developed libraries and models. Despite the direct support only for PyTorch, the adequacy of PyTorch-to-ONNX model conversion (despite its limitations in functionality) makes model interoperability between deep learning frameworks achievable.
        
        \textbf{Representing neural networks as syntax-tree code graphs}\quad We take advantage of the PyTorch "deep learning models as regular Python programs" design paradigm in order to represent neural network models in the form of static syntax-tree code graphs, rather than pre-defined neural component graphs. This unique approach allows us to support the full range of dynamic functionality, and arbitrary code execution available in normal PyTorch models. Our components simply encapsulate segments of normal PyTorch or Python code, and our visual designer implementation then allows for combining those elements into arbitrarily complex programs and neural network implementations. This method of code graph construction could hypothetically support any framework, although we have opted to tightly couple our system with PyTorch.
    
        \textbf{Hybrid visual and code-based approach}\quad To be useful, PrototypeML needs to prioritize visual clarity in the neural network design process, without reducing capability or model expressiveness. We concentrate on providing clear visual shortcuts and interfaces for specific elements of the neural network design process that contribute significant complexity or confusion (such as visualizing model architecture and data flow, skip connections, common matrix manipulation, etc.). For other aspects of the design process that prove notably more intuitive when written purely in code (such as complex custom layers), we opt not to force usage of visual elements that would negatively impact usability or functionality, but rather provide integrated code editors where appropriate, with auto-completion for visually defined parameters and variables. Our aim is not to replace all code writing in the development process, rather to allow for complex components encapsulating specific functionality to be coded once, easily debugged and maintained, and subsequently used in future models without requiring further coding.
        
    \subsection{Implementation details}
        PyTorch neural network models are implemented as regular Python classes that inherit from the PyTorch $nn.Module$ class, where each child $nn.Module$ class implements (1) an $init()$ function defining layers to be instantiated and their respective parameters, and (2) a $forward()$ function that defines the network data flow and computations to be performed at run-time. The following section details how we deconstruct the PyTorch class design paradigm into individual modular building blocks which can then be combined (through visual means) into complex models.
        
        \begin{figure}[H]
          \centering
          \includegraphics[width=\textwidth]{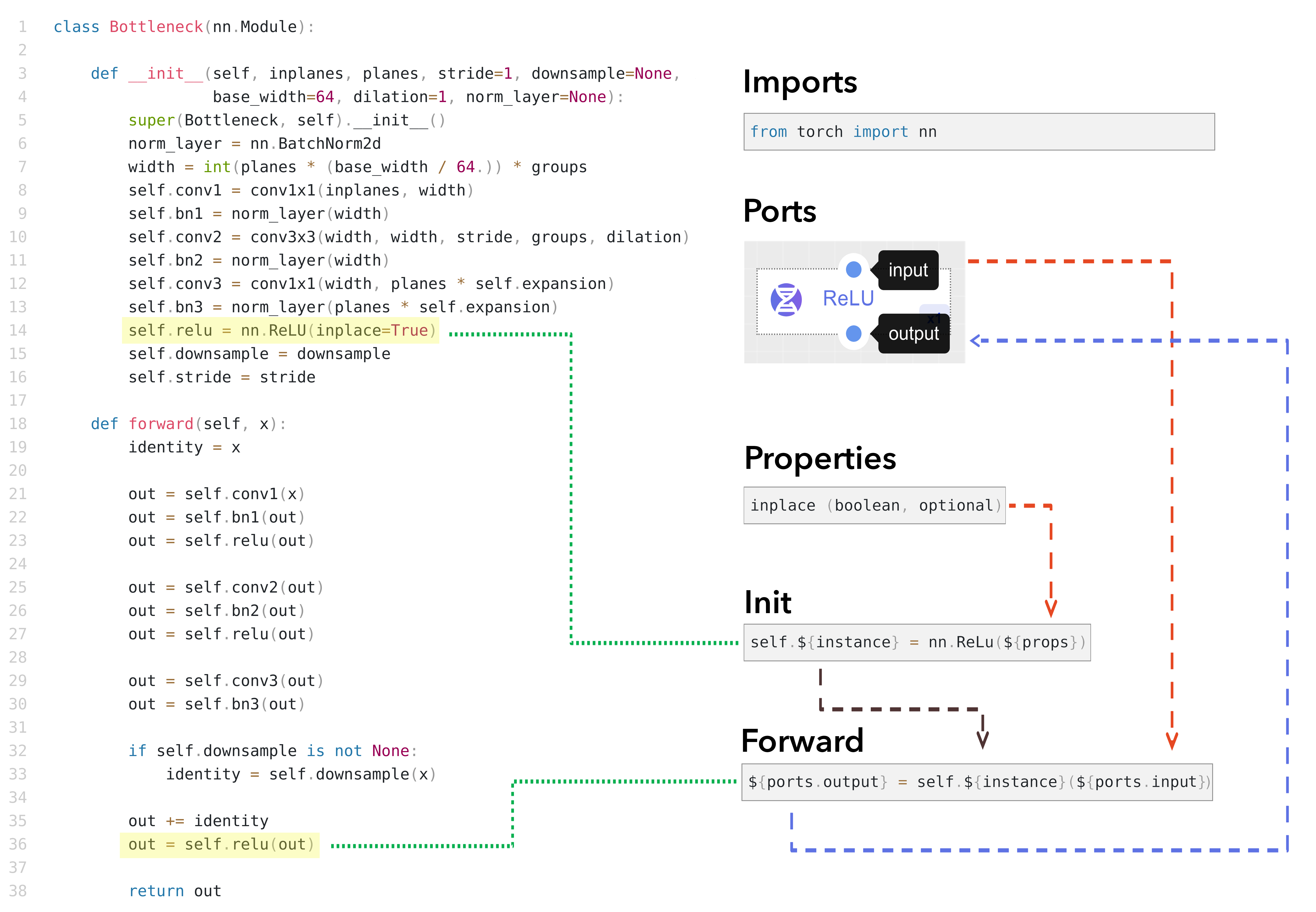}
          \caption{ReLU mutator implementation example}
          \label{fig:mutator}
        \end{figure}
    
        \textbf{Mutators}\quad The aim of the mutator is to wrap all the code necessary to initialize and execute a specific segment of functionality, such as a PyTorch layer, matrix operation, existing 3rd-party library or any arbitrary Python code using the PyTorch $init()$ and $forward()$ pattern. Each mutator is defined through a series of simple visual interfaces and built-in code editors. Specifically, mutators implement the following fields (special tokens identified by the $\$\{token\}$ syntax are replaced during code exporting with the appropriate values):
        \begin{enumerate}
            \item {[Optional]} Any imports required (such as the standard PyTorch library, \emph{or any 3rd-party PyTorch library or code base})
            \item {[Required]} The number of data inputs and outputs
            \item {[Optional]} The expected input shapes, and the method by which the output shapes can be calculated (for validation purposes)
            \item {[Optional]} Any parameters required by the component, and their validation logic (these values are made accessible within any of the component code editors via the $\$\{props\}$ object)
            \item {[Required]} The code to initialize the component within the $init()$ function
            \item {[Required]} The code to execute the component within the $forward()$ function
            \item {[Optional]} Any additional code required (e.g., a custom layer and backward calculation)
        \end{enumerate}
        
        Figure \ref{fig:mutator} demonstrates how a component in a standard PyTorch $nn.Module$ (in this case the built-in PyTorch ReLU activation) can be deconstructed into the various mutator fields. While we conceptually designed the mutator system to receive data via the defined inputs and outputs, and parameters via the parameter system, we leave it up to the component designer to choose the correct data-flow pattern for their components.
        \\
        
\begin{figure}[H]%
            \centering
            \subfigure[Parameter passing]{{\includegraphics[width=0.3\textwidth, keepaspectratio]{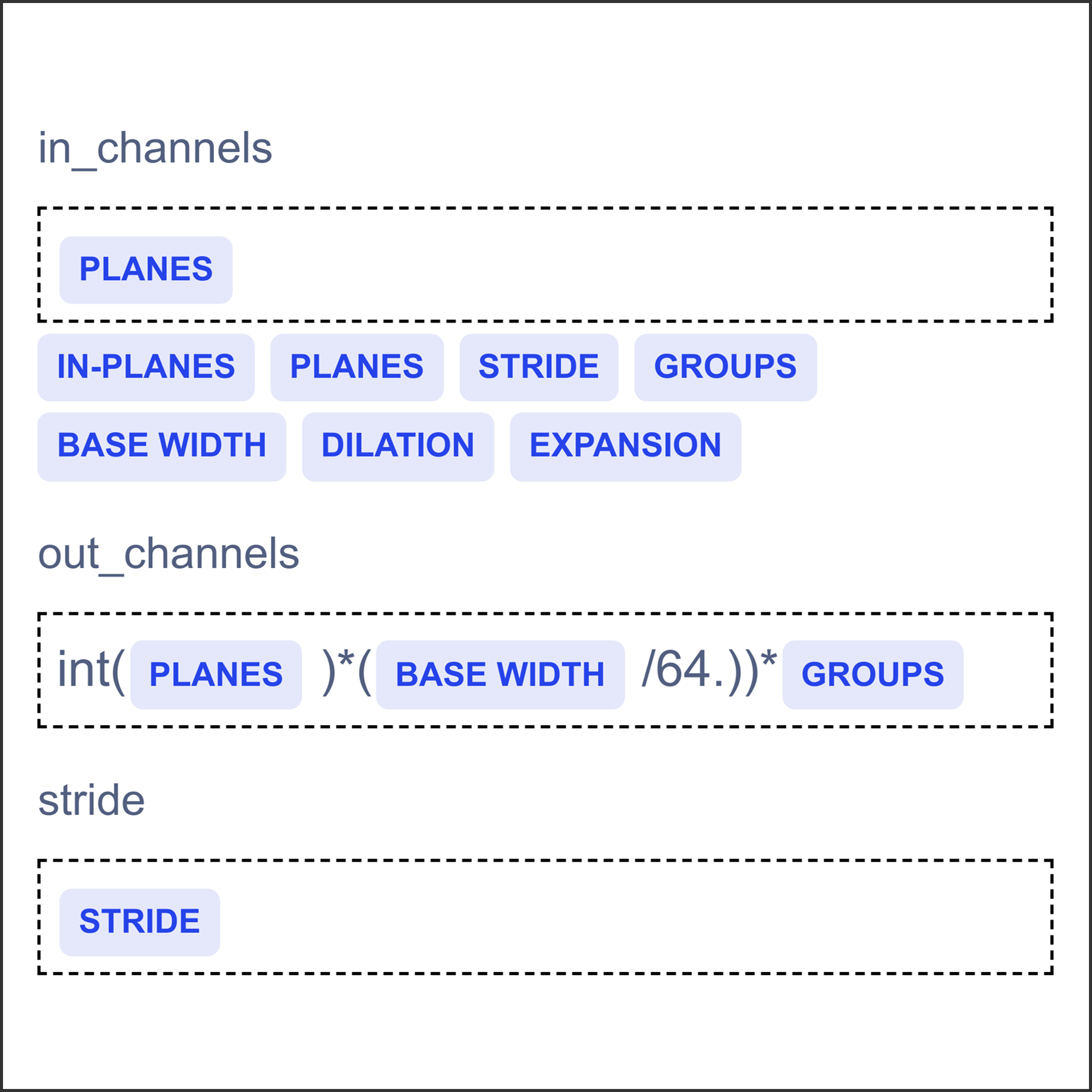} }\label{parameter_passing}}%
            \subfigure[Parameter validation]{{\includegraphics[width=0.3\textwidth, keepaspectratio]{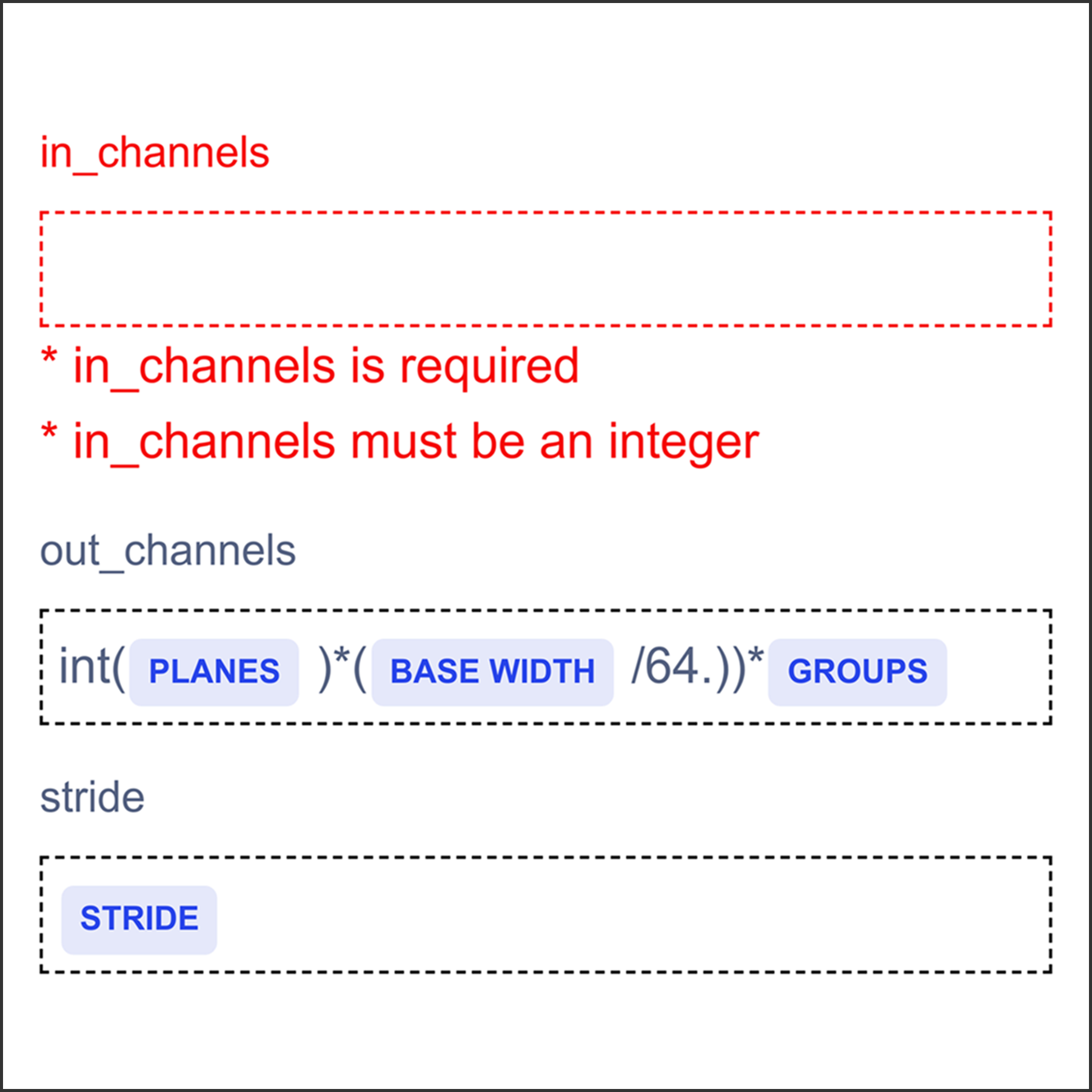} }\label{parameter_validation}}%
            \subfigure[Configurable tensor operations]{{\includegraphics[width=0.3\textwidth, keepaspectratio]{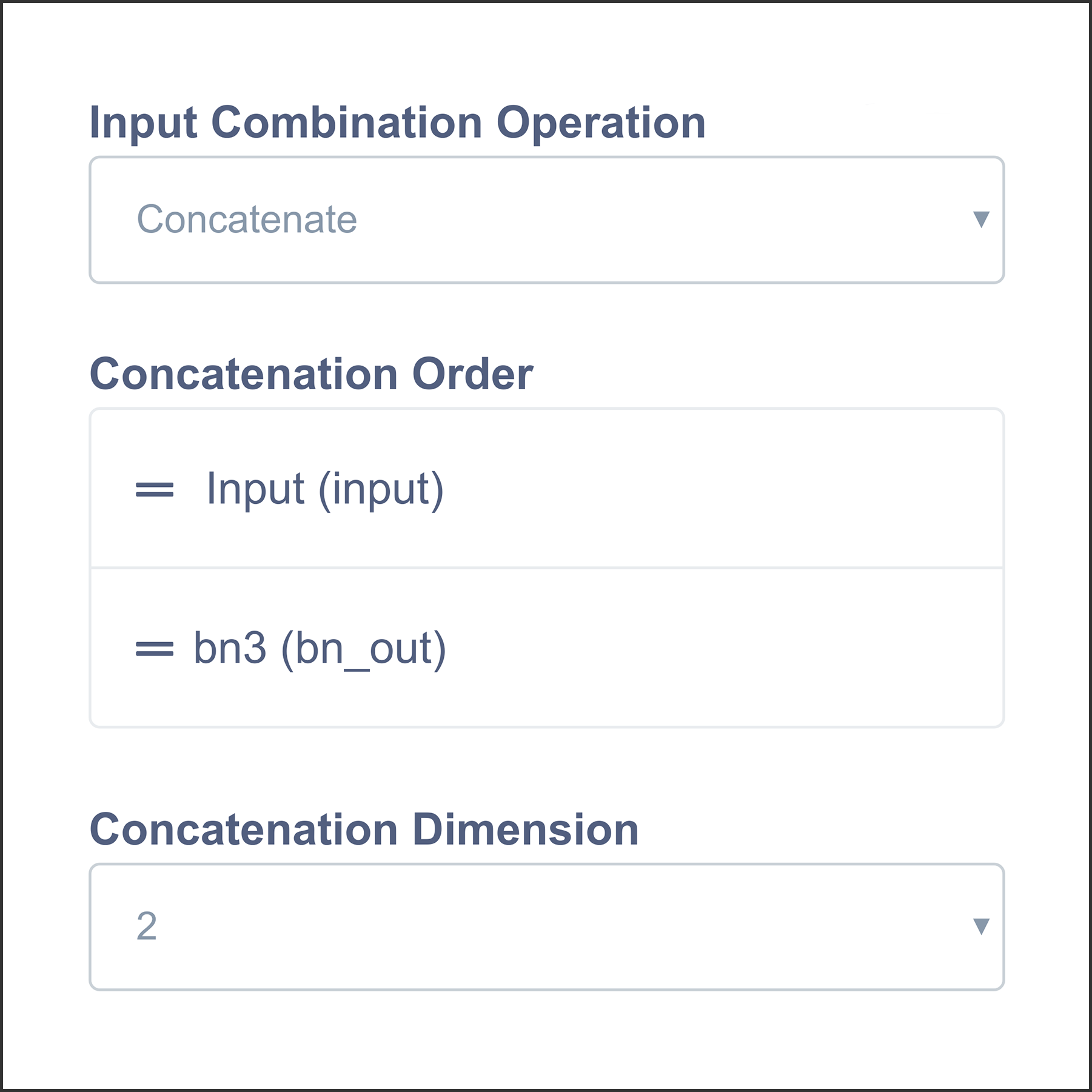} }\label{tensor_operations}}%
            \caption{Additional features}%
            \label{fig:additional features}%
        \end{figure}

        \textbf{Blocks}\quad Blocks act as a container format that encapsulates the relationships between child nodes. Each block implements (and is visualized by) a directed acyclic graph consisting of \{1, \dots, n\} user-placed instances of mutators or other blocks, structured between special \emph{Input} and \emph{Output} nodes. Similar to mutators, blocks also define a set of input and outputs, expected shapes and validation fields, and any parameters required by the block. Component instance parameters and options appear and are edited in the right sidebar (see figure \ref{interface3}), and are accessed by clicking upon the node within the block graph.  These parameters may be passed down to child nodes, and combined by defining block-level variables in order to encode complex functionality (see figure \ref{parameter_passing}).\\\\
        Through simple drag and drop interactions, component instances can be added to the block graph from the left sidebar (see figure \ref{interface1}), and edges between specific inputs and outputs on nodes can be defined (see figure \ref{interface2}). Complex neural network structures such as skip connections between nodes can also be easily expressed (see figure \ref{interface2}). We further simplify many frequently used design patterns by providing fast interfaces to features such as common activation functions, and desired behavior (add, concatenation, etc.) when two or more edges join at the same input (see figure \ref{tensor_operations}).
        \\
        
        \textbf{Loops and conditionals}\quad Loops and conditional logic are essential to support the full range of PyTorch module capabilities. In PrototypeML, complex looping behavior can be achieved by repeating mutator or block nodes within a block graph \{1, \dots, n\} or $\$\{parameter\}$ times via the parameter interface, as long as input and output shapes are compatible. At code generation time, the relevant node will be placed within a loop. Furthermore, repeated blocks and mutators may access a special $\$\{repeat\_index\}$ token that allows for complex logic based on loop iteration.\\\\
        Conditional logic is implemented within the block graph in the form of a special "if" control node, for which a conditional clause may be specified. The conditional control node mirrors inputs on both the left side (traversed if the condition is true), and right side (traversed if the condition is false) of the graph node, and thus provides a clear visual conditional data flow interface.\\
        
        \textbf{Model validation \& debugging}\quad While PyTorch models are easier to debug due to the Pythonic nature of their construction, they largely lack ahead-of-time design validation mechanisms and execution can halt in the middle of training, causing loss of progress. In order to improve the model debugging and validation experience, we implemented two forms of validation that provide end-users with a form of built-in documentation:
        \\\\
        (1) Validation logic can be easily applied to input parameter definitions when creating mutators and blocks. Parameter validation logic includes checking the presence of required parameters, type checking (int, float, matrix, etc.), and parameter value and shape checking (see figure \ref{parameter_validation}).\\\\
        (2) Mutator and block creators may also optionally specify how their output shapes are calculated based on the data inputs and parameters they defined (e.g., if a mutator adds a dimension to a data input, the output shape of the mutator would be the shape of the input data plus one dimension). If a block or mutator specifies their output shape calculation, they may also optionally specify the expected shape of their data inputs, which will then be validated against those entering the node from the previous component in the graph. Specifically, input shape validation and output shape calculation are optional. If a specific component does not provide these fields, the next block used in the graph which contains input shape validation will assume it received the correct shape and validation will continue further down the graph, simply skipping the validation-disabled component (which triggers a user warning).\\
        
    \subsection{Exporting code}
    We implement one-click PyTorch code generation that outputs human readable, easily debuggable and maintainable code on par with other neural network libraries, although the quality of the code exported from our platform largely depends upon the quality of the user-implemented mutators. Code generation performs a task similar to a linker concatenating different modules into program sections, in that it constructs PyTorch neural network $nn.Module$ classes by traversing the component code-graph
    \begin{algorithm}[H]
            \SetAlgoLined
            % \KwResult{Generated PyTorch code}
            \SetKwInOut{Input}{Input}
            \SetKwInOut{Output}{Output}
            \SetKwFunction{FGenerateCode}{GenerateCode}
            
            \Input{\{1, \dots, n\} $project\_blocks$ (all the blocks contained within a project)}
            \Output{One file for each $project\_block$ containing $nn.Module$ class code}
            
            \SetKwProg{Fn}{Def}{:}{}
            \Fn{\FGenerateCode{$project\_blocks$}}{
                $files$ = [ ]\\
                \For{$block$ in $project\_blocks$}{
                    \tcp{topologically sort block nodes from input node to output node and verify that the graph is directed and acyclic}
                    $sorted\_graph$ = topological\_sort$(block)$\\
                    \For{$node$ in $sorted\_graph$}{
                        construct node imports\\
                        construct node parameters\\
                        construct node data inputs\\
                        construct mutator and block init statements and connect parameters\\
                        construct mutator and block forward statements and connect inputs and parameters\\
                        construct node data outputs\\
                    }
                    $block\_code$ = FillModuleCodeTemplate($block.name$, $imports$, $parameters$,\\
                    \qquad\qquad\qquad\qquad\qquad\qquad\qquad\qquad\ $data\_inputs$, $data\_outputs$,\\
                    \qquad\qquad\qquad\qquad\qquad\qquad\qquad\qquad\ $inits$, $forwards$ )\\
                    $files$.append($block\_code$)
                }
                \KwRet $files$
            }
            \caption{Code generation process}
        \end{algorithm}
        defined by each block, combining block and mutator calls into the $init()$ and $forward()$ paradigm, and resolving input and output dependencies and parameters between the various components. Specifically, we implement this in two stages: (1) graph parsing and dependency construction and (2) code template filling.

\section{Flexibility and complexity abstraction}
    The PrototypeML mutator and block design paradigm provides the ability to quickly and easily implement complex and highly expressive modularized neural networks in a visually intuitive fashion.
    
\begin{figure}[H]
        \centering
        \subfigure[Resnet Block]{{\includegraphics[width=0.3\textwidth, keepaspectratio]{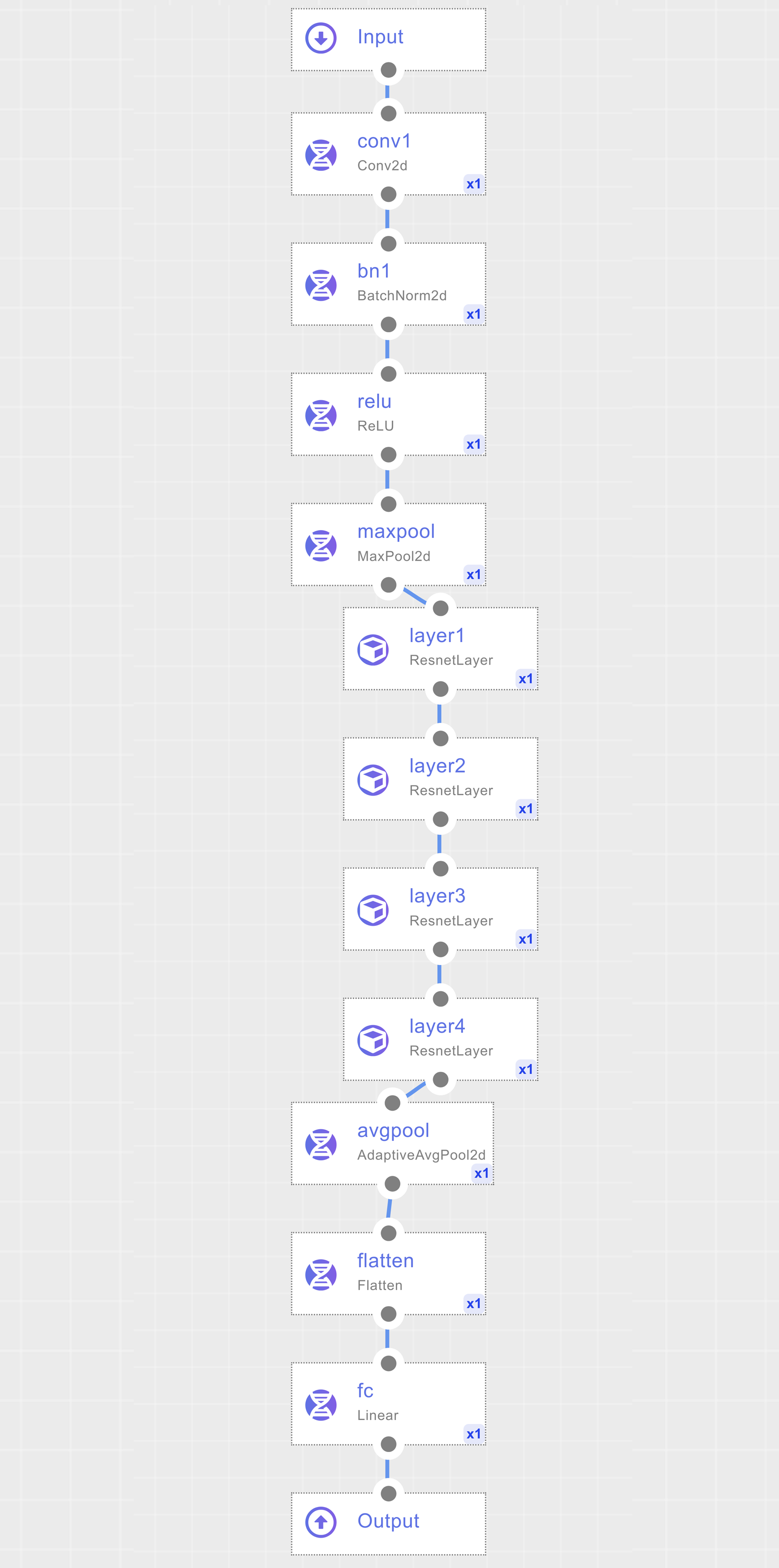} }\label{resnet_block}}%
        \subfigure[Resnet Layer Block]{{\includegraphics[width=0.3\textwidth, keepaspectratio]{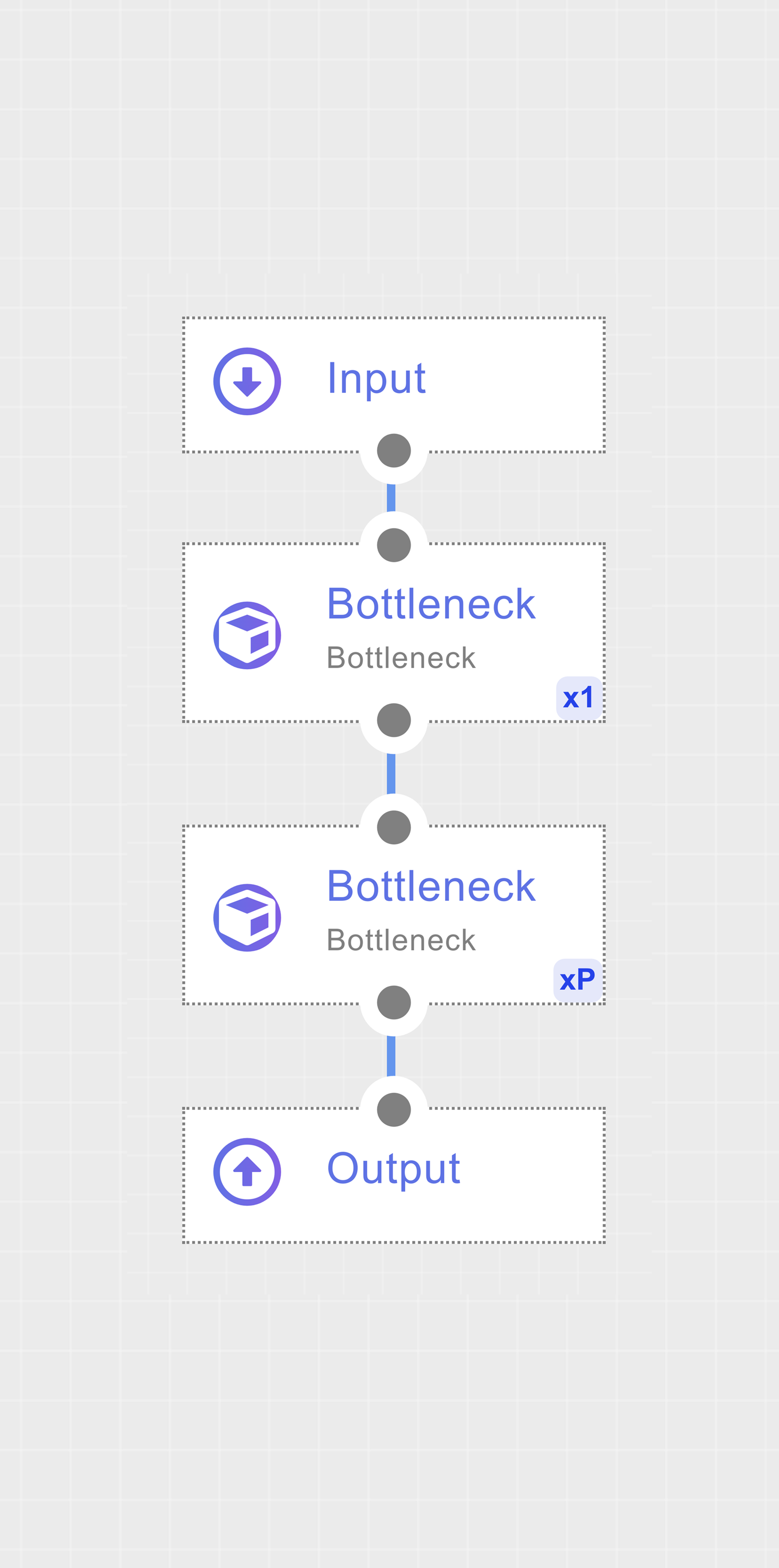} }\label{resnet_layer}}%
        \subfigure[Resnet Bottleneck Block]{{\includegraphics[width=0.3\textwidth, keepaspectratio]{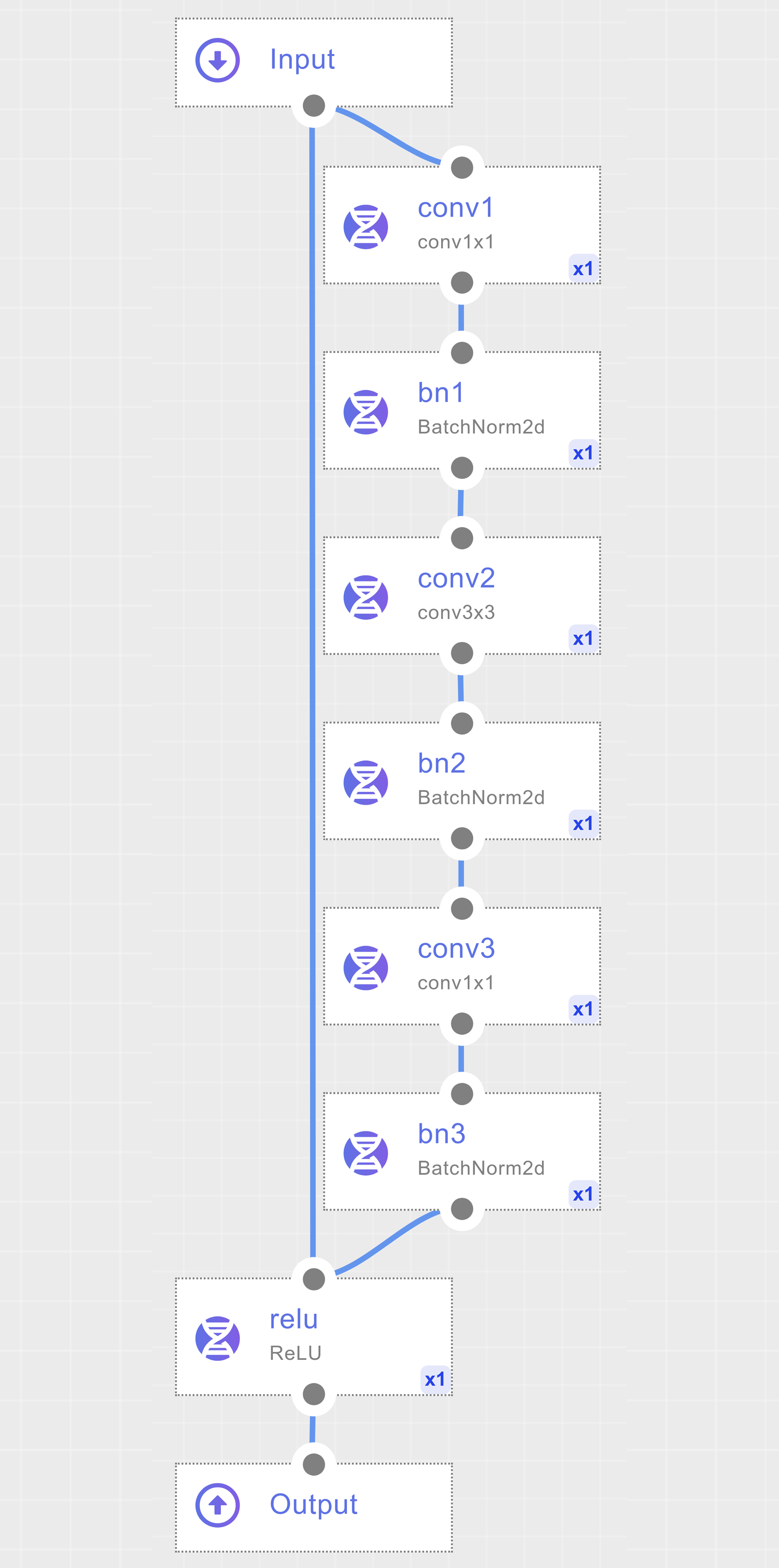} }\label{bottleneck}}%
        \caption{Resnet architecture \cite{resnet}}%
        \label{fig:resnet_architecture}%
    \end{figure}

    Figure \ref{fig:resnet_architecture} demonstrates how a Resnet architecture \cite{resnet} might be built within the platform. Specifically figure \ref{resnet_block} defines the Resnet architecture containing several $convolution$, $batch norm$, $pooling$ and $fully$-$connected$ mutators, in addition to $ResnetLayer$ blocks. Figure \ref{resnet_layer} defines the $ResnetLayer$ block, and demonstrates the usage of the $Bottleneck$ block which is repeated a parameter $P$ number of times according to the parameter definition of the four $ResnetLayers$. Figure \ref{bottleneck} defines the $Bottleneck$ block, and implements the skip connection between the input and the output of several convolutions.
    
    In addition to the flexibility and support for arbitrarily complex neural networks within the PrototypeML system, the abstractive nature of the mutator and block design paradigm allows for users of varying levels of expertise to quickly and easily build level-appropriate neural network designs without being forced to interface with more complicated components. For example, researchers might find themselves frequently implementing custom mutators and coded functionality, whereas an industry practitioner might combine pre-existing components into complex applications, and a new student to the field might only utilize pre-built architectures without delving deeper.
    
\section{Neural network social component repository}
    While there exists a plethora of third-party libraries written in PyTorch that implement a wide array of custom layers, architectures, and other deep learning functionality, these libraries are often monolithic, implement non-standard interfaces and usage patterns, and are generally not easily combined with other code bases. Taking advantage of the standardized modular format we implement for expressing individual components of neural networks and their usage patterns within the mutator and block paradigm, we propose and implement a system (similar to GitHub \cite{github}) that allows users to easily offer their blocks and mutators designed via the ProtypeML platform in a publicly accessible, open source, and centralized manner via a social neural network component repository. In addition to the components themselves, usage documentation may be provided, pre-trained component weights and scores on various datasets may be uploaded, and versioned component releases help to facilitate easy component maintenance, future updates and package management.
    \\\\
    Components available on the component repository can then be easily added to new projects via the design interface, allowing for complex networks to be created and combined in unique and powerful ways with very little overhead. This has applications for both industry in speeding up design and development of task-specific networks, and for the research community in expediting the exploration of new neural network design patterns and combinations.

\section{Future work}
    In addition to improving various aspects of the design process, we perceive several worthwhile directions in which to continue development of the PrototypeML platform.\\\\
    \textbf{Data loader \& training loop designer}\quad Currently data loaders and complex training loops are particularly time consuming to build. We believe complexity can be significantly reduced through the application of visual interfaces to these tasks.\\\\
    \textbf{Training \& experiment tracking}\quad For development continuity and ease-of-use purposes, the ability to train models and track experiments should be implemented directly within the tool. Specifically, we wish to offer local machine training, in addition to 3rd party cloud-based training methods.\\\\
    \textbf{Automatic output shape calculation for validation}\quad In order to validate input shapes, mutator creators must manually define the process by which their output shapes can be calculated. We hope to automate this process in the future by combining known shape manipulation operations to derive the output shape of arbitrary code, in a process modeled after the implementation of AutoGrad \cite{paszke2017automatic}).\\\\
    \textbf{Performance analysis of network graph and design feedback}\quad Due to the availability of a vast number of tactics that each incrementally improve neural network performance or accuracy, we plan to offer automated neural network design feedback and suggestions at appropriate times. These suggestions could be further augmented with suggestions offered by auto AI and meta learning solutions.\\\\
    \textbf{Data-driven network auto completion}\quad When implementing networks, we often follow specific design patterns (such as placing max pooling layers after convolutions). We are actively working on models that attempt to predict the next used layer based on the current graph in order to offer neural network auto-completion.\\\\
    \textbf{Opportunities for meta learning} Utilizing the social neural network standardized component repository and recorded usages of components as a form of tagged dataset, we believe that there exists the potential for advancement of meta-learning research initiatives. We plan on releasing a dataset of neural network component usage patterns in the future.\\\\
    \textbf{Visual learning / visual teaching}\quad Due to the visual nature of the platform and the accompanying ease of understanding, we believe that there exists significant potential in the use of the platform as a deep learning teaching tool. We are actively working on an introduction to deep learning course that involves students completing a series of automatically-graded visual exercises that concentrate on teaching neural network theory, rather than code.

\pagebreak

\section*{Broader impact}
The visual approach and relative ease of creating neural networks using our tool helps to lowers the deep learning barrier to entry. While we believe that this helps shift the emphasis from complex math and computer science prerequisites to prioritizing understanding of neural network theory and behavior, we recognize the importance of those computational skills to the continued advancement of deep learning, and acknowledge that by making them non-compulsory, fewer people might be motivated to pursue those vital sciences in depth.

However the partial decoupling of neural network design from significant prerequisite knowledge also presents an opportunity, specifically for industry practitioners. A looming significant societal concern is posed by the loss of human jobs as companies begin to apply greater numbers of deep learning solutions to their business practices. Specifically the McKinsey Global AI Survey \cite{mckinsey_survey} shows that in the coming years, companies hope to up-skill their existing employees to build deep learning solutions, rather than hire outside talent. We believe that many industry usages of deep learning can be applied without substantial computational background using our tool, thus reducing up-skill resource expenditure, and helping ensure that engineers retain their positions in the future.

Furthermore, we believe that the proposed publicly accessible neural network component repository allows for deep learning solutions to complex tasks to be developed with significantly less overhead, thus allowing specifically disadvantaged populations to more easily take advantage of the powerful capabilities offered by neural network solutions.

\bibliography{references}

\begin{thebibliography}{23}
\providecommand{\natexlab}[1]{#1}
\providecommand{\url}[1]{\texttt{#1}}
\expandafter\ifx\csname urlstyle\endcsname\relax
  \providecommand{\doi}[1]{doi: #1}\else
  \providecommand{\doi}{doi: \begingroup \urlstyle{rm}\Url}\fi

\bibitem[git()]{github}
Github: Build software better, together.
\newblock URL \url{https://github.com/}.

\bibitem[mck()]{mckinsey_survey}
Global ai survey: Ai proves its worth, but few scale impact.
\newblock URL
  \url{https://www.mckinsey.com/featured-insights/artificial-intelligence/global-ai-survey-ai-proves-its-worth-but-few-scale-impact}.

\bibitem[son()]{sony}
Neural network console.
\newblock URL \url{https://dl.sony.com/}.

\bibitem[nvi(2019)]{nvidia}
Nvidia digits, Jun 2019.
\newblock URL \url{https://developer.nvidia.com/digits}.

\bibitem[dee(2020)]{deepcognition}
Deep learning studio, May 2020.
\newblock URL \url{https://deepcognition.ai/features/deep-learning-studio/}.

\bibitem[Abadi et~al.(2015)Abadi, Agarwal, Barham, Brevdo, Chen, Citro,
  Corrado, Davis, Dean, Devin, Ghemawat, Goodfellow, Harp, Irving, Isard, Jia,
  Jozefowicz, Kaiser, Kudlur, Levenberg, Man\'{e}, Monga, Moore, Murray, Olah,
  Schuster, Shlens, Steiner, Sutskever, Talwar, Tucker, Vanhoucke, Vasudevan,
  Vi\'{e}gas, Vinyals, Warden, Wattenberg, Wicke, Yu, and Zheng]{tensorflow}
M.~Abadi, A.~Agarwal, P.~Barham, E.~Brevdo, Z.~Chen, C.~Citro, G.~S. Corrado,
  A.~Davis, J.~Dean, M.~Devin, S.~Ghemawat, I.~Goodfellow, A.~Harp, G.~Irving,
  M.~Isard, Y.~Jia, R.~Jozefowicz, L.~Kaiser, M.~Kudlur, J.~Levenberg,
  D.~Man\'{e}, R.~Monga, S.~Moore, D.~Murray, C.~Olah, M.~Schuster, J.~Shlens,
  B.~Steiner, I.~Sutskever, K.~Talwar, P.~Tucker, V.~Vanhoucke, V.~Vasudevan,
  F.~Vi\'{e}gas, O.~Vinyals, P.~Warden, M.~Wattenberg, M.~Wicke, Y.~Yu, and
  X.~Zheng.
\newblock {TensorFlow}: Large-scale machine learning on heterogeneous systems,
  2015.
\newblock URL \url{https://www.tensorflow.org/}.
\newblock Software available from tensorflow.org.

\bibitem[Bai et~al.(2019)Bai, Lu, Zhang, et~al.]{onnx}
J.~Bai, F.~Lu, K.~Zhang, et~al.
\newblock Onnx: Open neural network exchange.
\newblock \url{https://github.com/onnx/onnx}, 2019.

\bibitem[Chollet et~al.(2015)]{keras}
F.~Chollet et~al.
\newblock Keras.
\newblock \url{https://keras.io}, 2015.

\bibitem[Collobert et~al.(2002)Collobert, Bengio, and Marithoz]{torch}
R.~Collobert, S.~Bengio, and J.~Marithoz.
\newblock Torch: A modular machine learning software library.
\newblock 11 2002.

\bibitem[Garg et~al.(2018)Garg, Prabhu, Yadav, Ramrakhya, Agrawal, and
  Batra]{fabrik}
U.~Garg, V.~Prabhu, D.~Yadav, R.~Ramrakhya, H.~Agrawal, and D.~Batra.
\newblock Fabrik: An online collaborative neural network editor, 2018.

\bibitem[He et~al.(2015)He, Zhang, Ren, and Sun]{resnet}
K.~He, X.~Zhang, S.~Ren, and J.~Sun.
\newblock Deep residual learning for image recognition, 2015.

\bibitem[Howard et~al.(2018)]{fastai}
J.~Howard et~al.
\newblock Fastai.
\newblock \url{https://github.com/fastai/fastai}, 2018.

\bibitem[Jia et~al.(2014)Jia, Shelhamer, Donahue, Karayev, Long, Girshick,
  Guadarrama, and Darrell]{caffe}
Y.~Jia, E.~Shelhamer, J.~Donahue, S.~Karayev, J.~Long, R.~Girshick,
  S.~Guadarrama, and T.~Darrell.
\newblock Caffe: Convolutional architecture for fast feature embedding, 2014.

\bibitem[Lang et~al.(2019)Lang, Bravo-Marquez, Beckham, Hall, and
  Frank]{lang2019wekadeeplearning4j}
S.~Lang, F.~Bravo-Marquez, C.~Beckham, M.~Hall, and E.~Frank.
\newblock Wekadeeplearning4j: A deep learning package for weka based on
  deeplearning4j.
\newblock \emph{Knowledge-Based Systems}, 178:\penalty0 48 -- 50, 2019.
\newblock ISSN 0950-7051.
\newblock \doi{https://doi.org/10.1016/j.knosys.2019.04.013}.
\newblock URL
  \url{http://www.sciencedirect.com/science/article/pii/S0950705119301789}.

\bibitem[Neubig et~al.(2017)Neubig, Dyer, Goldberg, Matthews, Ammar,
  Anastasopoulos, Ballesteros, Chiang, Clothiaux, Cohn, Duh, Faruqui, Gan,
  Garrette, Ji, Kong, Kuncoro, Kumar, Malaviya, Michel, Oda, Richardson,
  Saphra, Swayamdipta, and Yin]{dynet}
G.~Neubig, C.~Dyer, Y.~Goldberg, A.~Matthews, W.~Ammar, A.~Anastasopoulos,
  M.~Ballesteros, D.~Chiang, D.~Clothiaux, T.~Cohn, K.~Duh, M.~Faruqui, C.~Gan,
  D.~Garrette, Y.~Ji, L.~Kong, A.~Kuncoro, G.~Kumar, C.~Malaviya, P.~Michel,
  Y.~Oda, M.~Richardson, N.~Saphra, S.~Swayamdipta, and P.~Yin.
\newblock Dynet: The dynamic neural network toolkit, 2017.

\bibitem[Paszke et~al.(2017)Paszke, Gross, Chintala, Chanan, Yang, DeVito, Lin,
  Desmaison, Antiga, and Lerer]{paszke2017automatic}
A.~Paszke, S.~Gross, S.~Chintala, G.~Chanan, E.~Yang, Z.~DeVito, Z.~Lin,
  A.~Desmaison, L.~Antiga, and A.~Lerer.
\newblock Automatic differentiation in pytorch.
\newblock 2017.

\bibitem[Paszke et~al.(2019)Paszke, Gross, Massa, Lerer, Bradbury, Chanan,
  Killeen, Lin, Gimelshein, Antiga, Desmaison, Kopf, Yang, DeVito, Raison,
  Tejani, Chilamkurthy, Steiner, Fang, Bai, and Chintala]{pytorch}
A.~Paszke, S.~Gross, F.~Massa, A.~Lerer, J.~Bradbury, G.~Chanan, T.~Killeen,
  Z.~Lin, N.~Gimelshein, L.~Antiga, A.~Desmaison, A.~Kopf, E.~Yang, Z.~DeVito,
  M.~Raison, A.~Tejani, S.~Chilamkurthy, B.~Steiner, L.~Fang, J.~Bai, and
  S.~Chintala.
\newblock Pytorch: An imperative style, high-performance deep learning library.
\newblock In H.~Wallach, H.~Larochelle, A.~Beygelzimer, F.~d\textquotesingle
  Alch\'{e}-Buc, E.~Fox, and R.~Garnett, editors, \emph{Advances in Neural
  Information Processing Systems 32}, pages 8024--8035. Curran Associates,
  Inc., 2019.
\newblock URL
  \url{http://papers.neurips.cc/paper/9015-pytorch-an-imperative-style-high-performance-deep-learning-library.pdf}.

\bibitem[Perrault et~al.(2019)Perrault, Shoham, Brynjolfsson, Clark,
  Etchemendy, Grosz, Lyons, Manyika, Mishra, Niebles, and
  et~al.]{ai_index_report}
R.~Perrault, Y.~Shoham, E.~Brynjolfsson, J.~Clark, J.~Etchemendy, B.~Grosz,
  T.~Lyons, J.~Manyika, S.~Mishra, J.~C. Niebles, and et~al.
\newblock \emph{The AI Index 2019 Annual Report}.
\newblock Dec 2019.
\newblock URL
  \url{https://hai.stanford.edu/sites/default/files/ai_index_2019_report.pdf}.

\bibitem[Roeder(2010)]{netron}
L.~Roeder.
\newblock Netron.
\newblock \url{https://github.com/lutzroeder/netron}, 2010.

\bibitem[Seide and Agarwal(2016)]{cntk}
F.~Seide and A.~Agarwal.
\newblock Cntk: Microsoft's open-source deep-learning toolkit.
\newblock pages 2135--2135, 08 2016.
\newblock \doi{10.1145/2939672.2945397}.

\bibitem[Tamilselvam et~al.(2019)Tamilselvam, Panwar, Khare, Aralikatte,
  Sankaran, and Mani]{ibm}
S.~Tamilselvam, N.~Panwar, S.~Khare, R.~Aralikatte, A.~Sankaran, and S.~Mani.
\newblock A visual programming paradigm for abstract deep learning model
  development, 2019.

\bibitem[{Theano Development Team}(2016)]{theano}
{Theano Development Team}.
\newblock {Theano: A {Python} framework for fast computation of mathematical
  expressions}.
\newblock \emph{arXiv e-prints}, abs/1605.02688, May 2016.
\newblock URL \url{http://arxiv.org/abs/1605.02688}.

\bibitem[Tokui et~al.(2019)Tokui, Okuta, Akiba, Niitani, Ogawa, Saito, Suzuki,
  Uenishi, Vogel, and Vincent]{chainer}
S.~Tokui, R.~Okuta, T.~Akiba, Y.~Niitani, T.~Ogawa, S.~Saito, S.~Suzuki,
  K.~Uenishi, B.~Vogel, and H.~Y. Vincent.
\newblock Chainer: A deep learning framework for accelerating the research
  cycle, 2019.

\end{thebibliography}

\end{document}